\documentclass[preprint,12pt]{elsarticle}

\usepackage{amssymb}
\usepackage{amsmath}
\usepackage{algorithmic}               
\usepackage{algorithm}                 
\usepackage{amssymb}                   
\usepackage{multirow}                  
\usepackage{booktabs}                  
\usepackage{caption}                   
\usepackage{graphicx}                  
\usepackage{subfig}                    
\usepackage{tabularx}                  
\usepackage{array}                     

\journal{Nuclear Physics B}

\begin{document}

\begin{frontmatter}



\title{What Will Happen Next: Large Models-Driven Deduction for Emergency Instances}




\author[inst1]{Zhengqing Hu}
\author[inst1]{Dong Chen\cormark[1]}
\author[inst2]{Junkun Yuan}
\author[inst1]{Liang Liu}
\author[inst1]{Hua Wang}
\author[inst1]{Zhao Jin}
\author[inst2]{Yingchaojie Feng}
\author[inst2]{Wei Chen}
\author[inst1]{Mingliang Xu\cormark[1]}

\affiliation[inst1]{organization={Zhengzhou University},
            city={Zhengzhou},
            postcode={450001}, 
            state={Henan},
            country={China}}
\affiliation[inst2]{organization={Zhejiang University},
            city={Hangzhou},
            postcode={310058}, 
            state={Zhejiang},
            country={China}}

\ead{chendongai@zzu.edu.cn, iexumingliang@zzu.edu.cn}

\cortext[1]{Corresponding author. E-mail: chendongai@zzu.edu.cn, iexumingliang@zzu.edu.cn}

\begin{abstract}
Traditional simulation methods reproduce occurred emergency instances thr-
ough presetting to assist people in risk assessment and emergency decision-making. However, due to the lack of randomness and diversity, existing simulation systems struggle to fully explore the potential risk as emergency instances are scarce. In contrast, Large Models (LMs) can dynamically adjust generation strategies to introduce controllable randomness, while also possessing extensive prior knowledge and cross-domain knowledge transfer capabilities. Inspired by it, we propose the LMs-driven World Line Divergence System (WLDS), which enables diversified visualization and deduction of emergency instances in different domains. WLDS leverages LMs to deduce emergency instances in various development directions, and introduces the factual calibration and logical calibration mechanism to ensure factual accuracy and logical rigor during the deduction process. The interactive module can independently select deduction directions to avoid potential hallucinations that are difficult for the system to identify. Furthermore, by introducing the visualization module, WLDS forms simulation and deduction that combine text and images, which enhances interpretability. Extensive experiments conducted on the proposed Emergency Instances Deduction (EID) benchmark dataset demonstrate that WLDS achieves high-precision and high-fidelity simulation and deduction of emergency instances in multiple specific domains. Relevant experiments further demonstrate that WLDS can generate more emergency instances deduction data for users and provide support for better decision-making in similar emergency instances in the future.
\end{abstract}

\begin{graphicalabstract}
\includegraphics[width=\linewidth]{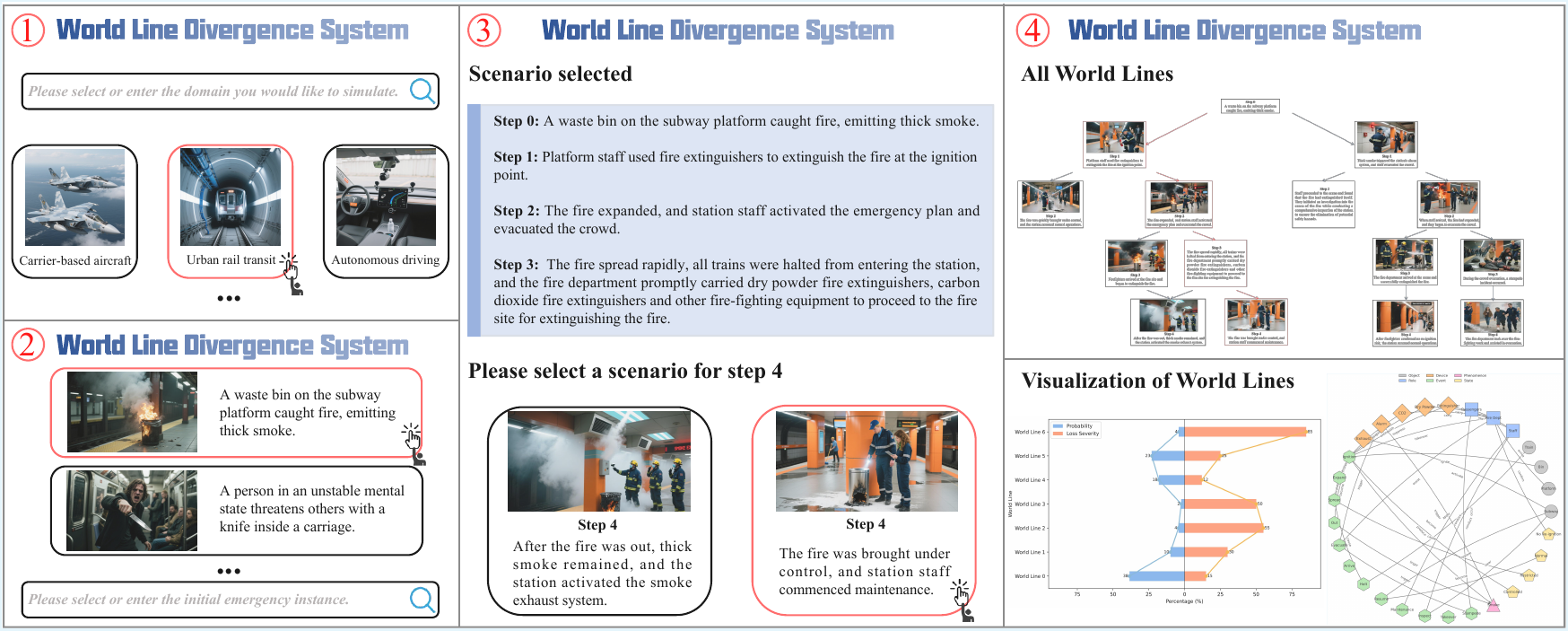}
\end{graphicalabstract}

\begin{highlights}
\item WLDS: a simulation and deduction system for emergency instances in few-shot, multi-domain professional settings.
\item Text–image fused, user-steerable interactive deduction with multi-branch world lines.
\item EID benchmark for emergency deduction, covering 10 domains and 4,300 three-step branched samples with expert labels.
\item Superior factual and logical consistency and higher scenario prediction accuracy, corroborated by expert evaluations.
\end{highlights}


\end{frontmatter}



\section{Introduction}
\label{sec1}
Digital simulation technologies against real-world scenarios greatly facilitate the understanding and reproduction of the processes and logic underlying the evolution of real events \cite{Santos04052022}.
By simulating real-world scenarios and deducing event development processes, simulation systems not only provide training materials for operators but also offer risk assessment basis for decision-makers \cite{10.1007/978-3-030-48021-9_37}.
Therefore, their performance directly determines the prediction accuracy of risks in the scenario and the effectiveness of emergency decision-making \cite{10989526}.

Existing simulation technologies have already demonstrated certain effectiveness in simulation modeling for normal scenarios, such as crowd simulation \cite{bae2025continuouslocomotivecrowdbehavior}.
However, they lack the ability to simulate and deduce emergency instances \cite{10611555,9721289}, as traditional simulation technologies suffer from the following problems:
(1) \textbf{Lack of randomness:} Simulation technologies against real-world scenarios can achieve digital mapping of physical entities.
However, in terms of logic deduction and state evolution, they overly rely on preset rules and lack the ability to model the randomness of event states and the diversity of event development paths in the physical world. 
(2) \textbf{Lack of diversity:} Specific domains such as autonomous driving and urban rail transit are characterized by high potential risks, rare but severe emergencies. 
For example, in urban rail transit scenarios, although fires are infrequent, they may lead to severe consequences such as traffic paralysis and stampedes.
Such emergency instances are crucial for improving the accuracy of simulation and deduction.
Due to the lack of relevant emergency instances, existing simulation and deduction technologies are ineffective and suffer from problems such as deviation from facts and illogical deduction.

\begin{figure}[!t]
\centering
\includegraphics[width=\linewidth]{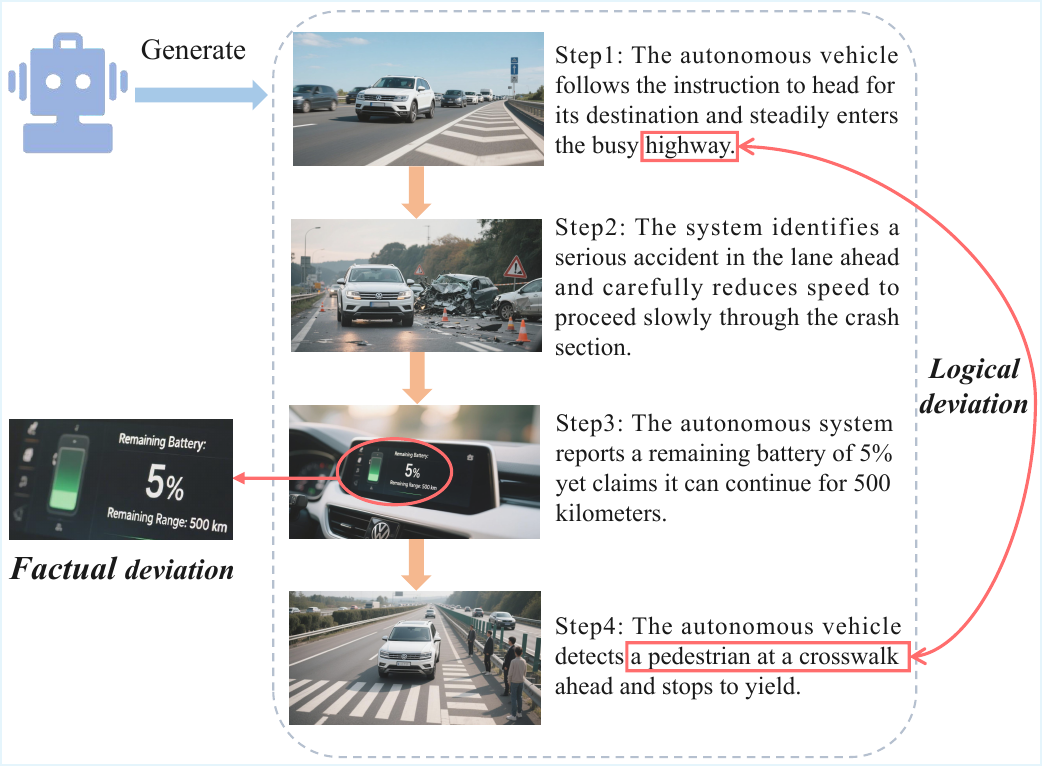}
\caption{Directly use LMs to simulate and deduce the process of autonomous driving. It includes two types of hallucination issues: factual deviation and logical deviation.}
\label{fig_2}
\end{figure}

In recent years, some studies have used Large Models (LMs) to dynamically adjust generation strategies and introduce controllable randomness to break the limitation of single scenario evolution patterns in traditional rule-driven simulations \cite{10801328, CHENG2024120238, 10.1145/3613904.3642024}.
For example, Li et al. \cite{10770822} designed the ChatSUMO system which combines LMs with the traffic simulation platform SUMO. It achieves the full-process automation from natural language input to urban-level traffic scenario generation and supports customized operations such as traffic signal optimization and vehicle path adjustment.
However, these studies lack the simulation of emergency instances.
Figure.\ref{fig_2} shows the result of directly using LMs to simulate and deduce the process of autonomous driving.
We summarize that directly using LMs to simulate and deduce is prone to the following two types of hallucination issues:
\textbf{Factual deviation:} The generated content that violates physical laws or domain specifications affects severely the reliability of the simulation and deduction.
As shown in Figure.\ref{fig_2}, in Step 3, the autonomous driving system reports only 5\% remaining battery, yet still claims that the vehicle can continue driving for 500 kilometers, which is physically impossible.
\textbf{Logical deviation:} During the simulation and deduction process, there are logical flaws such as causal disconnection and broken element consistency, which lead to the lack of logical rigor.
As shown in Figure.\ref{fig_2}, in Step 1, the vehicle is already on a busy highway, but in Step 4, LMs deduce that the same vehicle encounters a pedestrian crosswalk that should not exist on the highways.
Furthermore, due to the extensive prior knowledge and cross-domain knowledge transfer capabilities of LMs \cite{yuan2023hap, yuan2023domain, 10113208, 10.1145/3503161.3548059}, they can be used to migrate event knowledge of emergency instances from other domains to the target domain, thereby alleviating the problem of emergency instances scarcity \cite{chen2024logicdistillationlearningcode, chen2025kkaimprovingvisionanomaly, Chen_Zhuang_Zhang_Liu_Dong_Tang_2024, chen2024improvinglargemodelssmall}.

The world lines represent the trajectory of an object or event in spacetime \cite{hawking2large}.
Inspired by this concept, we propose \textbf{LMs-Driven \underline{W}orld \underline{L}ine \underline{D}ivergence \underline{S}ystem (WLDS)}, which aims to achieve high-precision and high-fidelity simulation and deduction of emergency instances, thereby providing references for safety assessment and decision support.
WLDS leverages the cross-domain transfer capability of LMs to migrate emergency instances knowledge from other domains to the target domain, thereby generating initial emergency instances. 
Subsequently, starting from the initial instance, WLDS uses LMs to generate multiple world lines with different development directions and allows users to independently select their desired deduction direction.
Meanwhile, WLDS introduces the dual calibration mechanism: 
The factual calibration mechanism achieves dynamic alignment between generated content and domain facts through real-time knowledge retrieval to ensure that each world line possesses factual reliability.
The logical calibration mechanism uses the logical discriminator to dynamically evaluate whether the logic between the current event and previous content is consistent, thereby ensuring that each world line has rigorous internal logic.
To address the deficiencies of existing evaluation systems, we propose an automated evaluation mechanism that quantitatively evaluates the performance of  WLDS by using factual consistency and logical consistency.
Moreover, we construct the Emergency Instances Deduction (EID) benchmark dataset to facilitate dynamic modeling and evaluation of emergency instances deduction. 
It consists of 10 sub-datasets which cover a wide range of domains from urban rail transit to autonomous driving and others.
The experimental results show that, compared to the baseline model, WLDS achieves a 7.08\% improvement in factual consistency and a 8.34\% improvement in logical consistency in the urban rail transit domain.
In the EID-Chemical plant sub-dataset, the scenarios prediction accuracy of WLDS is 8.50\% higher than that of the baseline model. 
Additionally, in the autonomous driving domain, WLDS received a high rating of 4.8 points from the domain experts.

The main contributions of this paper can be summarized as follows:  
\begin{enumerate}
    \item We systematically analyze the problems of existing simulation systems in the simulation and deduction of emergency instances, and discuss the two types of hallucination issues (factual deviation and logical deviation) of LMs in this field.  
    \item We propose LMs-Driven World Line Divergence System (WLDS), which combines with factual calibration and logical calibration mechanisms to achieve high-precision and high-fidelity simulation and deduction of emergency instances through LMs.  
    \item We construct the EID benchmark dataset, which consists of 10 sub-datasets with a total of 4300 data entries. This dataset can provide high-quality data support for optimizing and evaluating models for emergency instances deduction.
    \item We design an automated evaluation mechanism based on factual consistency and logical consistency. Extensive experiments have been conducted to demonstrate the effectiveness of WLDS in simulating and deducing emergency instances.
\end{enumerate}

\section{Related work}

\subsection{Scenario Generation Technology}
Scenario generation is a key supporting technology for complex environment modeling \cite{10529537, LU2025105613, 10693608}, and existing methods can be broadly classified into model-based and data-driven approaches \cite{10588675, niu2024planningsimulationmotionplanning}.

Model-based methods can generate continuous scenarios through mathematical modeling or rule systems. For example, DiffScene \cite{Xu_Petiushko_Zhao_Li_2025} employs diffusion models combined with adversarial optimization to produce high-quality safety-critical scenarios. Bagschik et al. \cite{8500632} proposed an ontology-based highway scenario generation method. Li et al. \cite{9662987} introduced a biologically inspired approach involving the exchange and mutation of scenario elements. 

Data-driven methods rely on large-scale scenario datasets, reproducing scenario characteristics and distributions by mining implicit information in the data \cite{10493025}. For example, Thal et al. \cite{9827198} generated high-coverage test cases based on real driving data. Bäumler et al. \cite{10347221} fused accident data with video-based traffic observations to produce more representative test scenarios. 

However, model-based methods are constrained by preset rules, and data-driven methods are restricted by the original data distribution, making it difficult to generate emergency instances deduction beyond existing patterns. 
To address the problem, WLDS introduces controllable randomness through LMs, dynamically adjusting generation strategies to enhance the diversity and randomness of emergency instances deduction.

\subsection{Simulation and Deduction Technology}
Simulation and deduction are core technologies for risk assessment and decision-making \cite{DEPAULAFERREIRA2020106868,Mourtzis02042020}. Traditional simulation models can integrate multiple perspectives to support complex decision-making \cite{BAUDRY2018257,1574544}, but their static nature limits applicability to dynamic scenarios.  

Digital twin technology, a key component of Industry 4.0 \cite{computers13040100}, continuously synchronizes with the physical system through real-time multi-source data \cite{DEAZAMBUJA202425,10.1145/3652620.3688253,su16083212,designs8060105}. For instance, Padovano et al. \cite{PADOVANO2024109839} combined BIM and sensor data to build pedestrian flow simulations and used LSTM to predict congestion, triggering automated alerts that reduced emergency response time by 40\%.  

Nevertheless, existing studies predominantly focus on the deduction of normal instances and lack research on emergency instances.
Due to the scarcity of emergency instances, models struggle to learn the unique evolutionary patterns of them, which leads to biases in prediction and poor emergency decision-making effectiveness.
WLDS alleviates it by transferring the knowledge of emergency instances from other domains to the target domain to support emergency instances deduction.

\subsection{LMs-Driven Simulation Technology}
In recent years, traditional platforms such as RLBench \cite{9001253} and CALVIN \cite{9788026} rely on manual design or simple randomization, which cannot meet the demands of complex tasks.
Leveraging their powerful semantic understanding and cross-modal reasoning capabilities, LMs provide a new impetus for advancing simulation technologies \cite{samak2025digitaltwinsmeetlarge, zhao2025surveylargelanguagemodels}.

Recent research has explored integrating LMs with simulation \cite{zhao2025sacascenarioawarecollisionavoidance,gan2025casebasedreasoningaugmentedlarge,zhang2024chatsceneknowledgeenabledsafetycriticalscenario}. Grutopia \cite{wang2024grutopiadreamgeneralrobots} constructs object–spatial relationship graphs for large-scale indoor scenario generation. RoboCasa \cite{nasiriany2024robocasalargescalesimulationeveryday} incorporates human demonstrations to optimize scenario layouts. LLMScenario \cite{10529537} employs prompt engineering and evaluation–feedback tuning to expand extreme cases in natural driving scenarios.  

However, in highly specialized domains such as autonomous driving and urban rail transit, LMs are prone to factual deviation and logical deviation. 
To address this issue, WLDS introduces the dual calibration mechanism to ensure that generated content adheres to physical laws and maintains logical rigor while preserving diversity.

\section{Method}  
\begin{figure*}[!t]
\centering
\includegraphics[width=\linewidth]{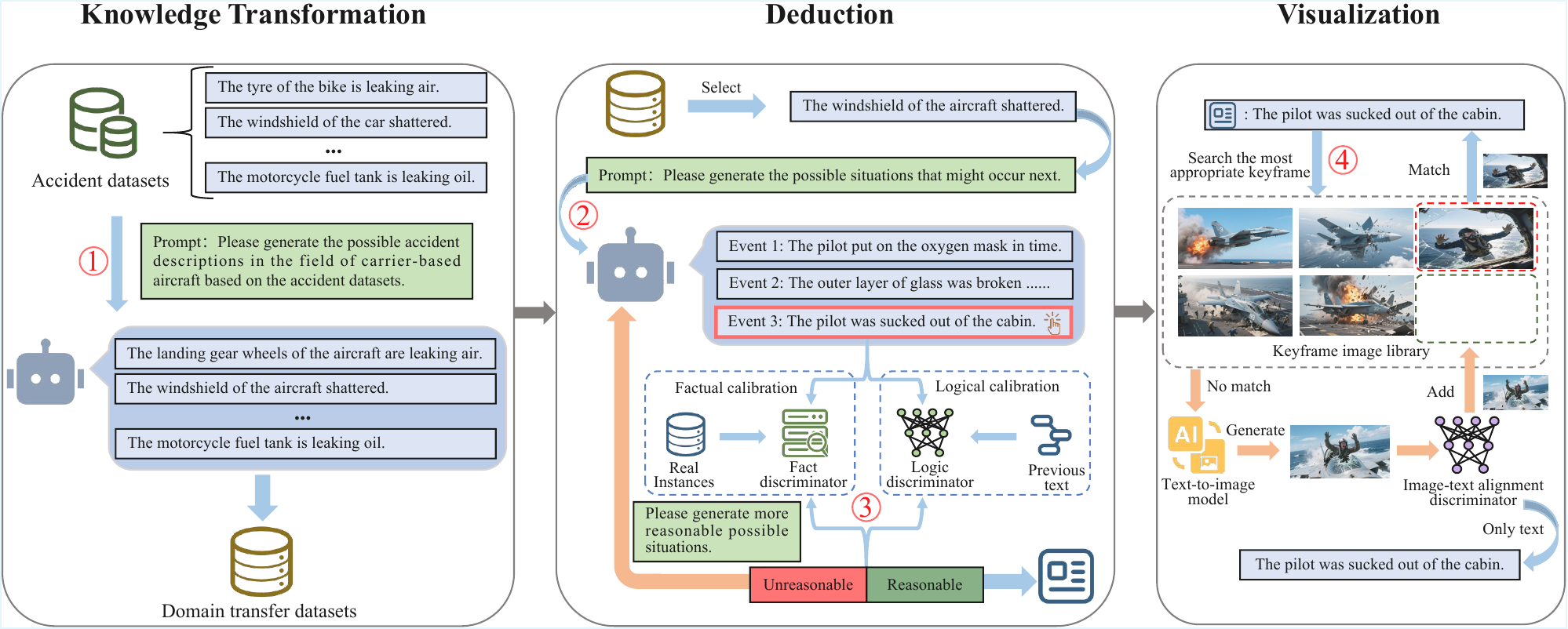}
\caption{The framework of the proposed WLDS. Step 1: WLDS uses LMs to generate the initial emergency instance for the target domain based on knowledge bases from other domains. Step 2: WLDS takes the generated emergency instances as the initial instance using LMs to generate descriptions of multiple potential scenarios. Step 3: For the potential scenarios, WLDS introduces the dual calibration mechanism to address factual deviation and logical deviation, and then users can select one scenario as the direction for deduction. Step 4: WLDS matches the corrected text with a keyframe image library. If matching fails, it will invoke a text-to-image model to generate images, and filter out images with significant semantic deviations via a text-image alignment discriminator.}
\label{fig_3}
\end{figure*}

As shown in Figure.\ref{fig_3}, we propose WLDS which can achieve high-precision and high-fidelity simulation and deduction of emergency instances.  
It comprises four core steps: emergency instances knowledge transformation, world line deduction, world line calibration, and world line visualization.  
In this section, we will further elaborate on the implementation process of the proposed WLDS.

\subsection{Knowledge Transformation}
To address the scarcity of emergency instances in specialized domains, we leverage the cross-domain knowledge transfer capability of LMs. We first collect emergency instances from different domains to construct an accident dataset $\mathcal{D}_{\mathrm{acc}}$, which serves as the knowledge base for transfer. Given a target domain $B$, WLDS employs the LMs to generate domain-specific emergency instances under a tailored prompt.

The generation prompt is as follows:  
\textit{``Please generate possible emergency instances descriptions of the domain $B$ based on the accident dataset.''} The generation process can be formalized as:
\begin{equation}
    e_B \sim p_{\theta}\big(\cdot \mid \mathcal{D}_{\mathrm{acc}}, \mathcal{K}_B, \pi \big).
\label{Eq 1}
\end{equation}
where $e_B$ denotes an emergency instance in the target domain $B$, $p_{\theta}$ is the conditional probability distribution defined by the LMs with parameters $\theta$, $\mathcal{K}_B$ represents the domain-specific knowledge base, and $\pi$ denotes the prompt.

By iteratively applying this process, we obtain a set of $N$ generated domain-specific instances: $\mathcal{D}_{\mathrm{trans}} = \{ e_1, e_2, \dots, e_N \}$.
This transferred dataset $\mathcal{D}_{\mathrm{trans}}$ provides the foundation for subsequent world line deduction.

\subsection{World Line Deduction}
To generate diverse world lines, WLDS introduces \textit{controlled randomness} into the LMs generation process. This randomness is governed by a \textit{temperature parameter} $\tau_k$, which balances logical plausibility with deduction diversity. Given an initial event $s_0 \in \mathcal{D}_{\mathrm{trans}}$, the LMs generate multiple possible subsequent scenarios under a domain-specific prompt:
\begin{equation}
    s_k \sim p_{\theta, \tau_k}\big(\cdot \mid s_0, \mathcal{K}_B\big), \quad k=1,2,\dots,M
\label{Eq 2}
\end{equation}
where $\mathcal{K}_B$ is the target domain knowledge base, and $\tau_k>0$ controls the diversity of generation. A larger $\tau_k$ yields more divergent scenarios, while a smaller $\tau_k$ leads to more deterministic outputs.

The randomness introduced by the temperature parameter $\tau_k$ can be formalized as:
\begin{equation}
    p_{\theta,\tau_k}(w_i \mid h) = \frac{\exp\left( z_i(h) / \tau_k \right)}{\sum_j \exp\left( z_j(h) / \tau_k \right)}.
\label{Eq 3}
\end{equation}
where $w_i$ is the $i$-th candidate token, $h$ denotes the current context, and $z_i(h)$ is the unnormalized logit for token $i$.

All generated scenarios form $\mathcal{S} = \{ s_1, s_2, \dots, s_M \}$.
The user then selects a scenario $s_{\mathrm{sel}} \in \mathcal{S}$ to form the initial world line: $W = [\,s_0,\, s_{\mathrm{sel}}\,]$.
This step provides a diverse foundation for subsequent calibration.

\subsection{World Line Calibration}
In multi-step deducing, event sequences generated by the model are prone to deviations caused by insufficient knowledge coverage or broken logical chains, which may lead to results that violate physical laws or domain-specific common sense.  
To address this, WLDS incorporates the \textit{dual calibration mechanism} after the initial construction of the world line, improving both its reliability and interpretability from two complementary perspectives: factual calibration and logical calibration.

(1) Factual Calibration: It focuses on ensuring the consistency between individual events and the domain knowledge base $\mathcal{K}_B$.  
For each event $s$, the system retrieves the most relevant fact $f(s)$ from $\mathcal{K}_B$ and computes a factual consistency score $\phi_{\mathrm{fact}}(s, f(s)) \in [0,1]$.  
If $\phi_{\mathrm{fact}}(s, f(s)) < \delta_{\mathrm{fact}}$, 
the event is revised under factual constraints:
\begin{equation}
    s^{\prime} \sim p_{\theta}\big(\cdot \mid s, f(s), \mathcal{K}_B\big).
\label{Eq 4}
\end{equation}
where $\delta_{\mathrm{fact}}$ is the factual consistency threshold.  
This step corrects explicit factual errors and ensures that each event is well-grounded in the underlying knowledge base.

(2) Logical Calibration: It targets the causal and temporal relationships between consecutive events.  
For each adjacent pair $(s_i, s_j)$, a logical consistency function $\psi_{\mathrm{logic}}(s_i, s_j) \in \{\text{valid}, \text{invalid}\}$ is applied.  
If $\psi_{\mathrm{logic}}(s_i, s_j) = \text{invalid}$, 
the subsequent event is regenerated with logic calibration:
\begin{equation}
    s^{\prime\prime} \sim p_{\theta}\big(\cdot \mid s_i, \mathcal{K}_B, \text{logic\_fix}\big).
\label{Eq 5}
\end{equation}
This mechanism mitigates accumulated reasoning errors, avoiding illogical jumps or contradictions in the world line.

(3) World Line Update: After both factual and logical calibration, the updated world line is:
\begin{equation}
    W^{*} = [\, s_0,\, s_{\mathrm{calibrated}}\,].
\label{Eq 6}
\end{equation}
where $s_{\mathrm{calibrated}}$ satisfies both factual and logical constraints.  

By combining these two forms of calibration, WLDS significantly mitigates the potential hallucination issue that LMs may exhibit, producing world lines that are both factually accurate and logically coherent.

\subsection{World Line Visualization}
To make the deduced world line more interpretable and accessible, WLDS incorporates a text-image integrated visualization mechanism.  
This mechanism not only generates keyframe images that are highly aligned with the semantics of each event, but also enhances the user’s perception and understanding of scenario evolution through multimodal fusion.

First, a keyframe image library is constructed: $\mathcal{I} = \{ I_1, I_2, \dots, I_P \}$, 
and a text-image alignment function $\alpha(s, I) \in [0,1]$ with a matching threshold $\delta_{\mathrm{align}}$ is defined. For each event $s$, the maximum alignment score is computed:
\begin{equation}
    \alpha_{\max}(s) = \max_{I \in \mathcal{I}} \alpha(s, I),
\label{Eq 7}
\end{equation}

\begin{figure*}[!t]
\centering
\includegraphics[width=\linewidth]{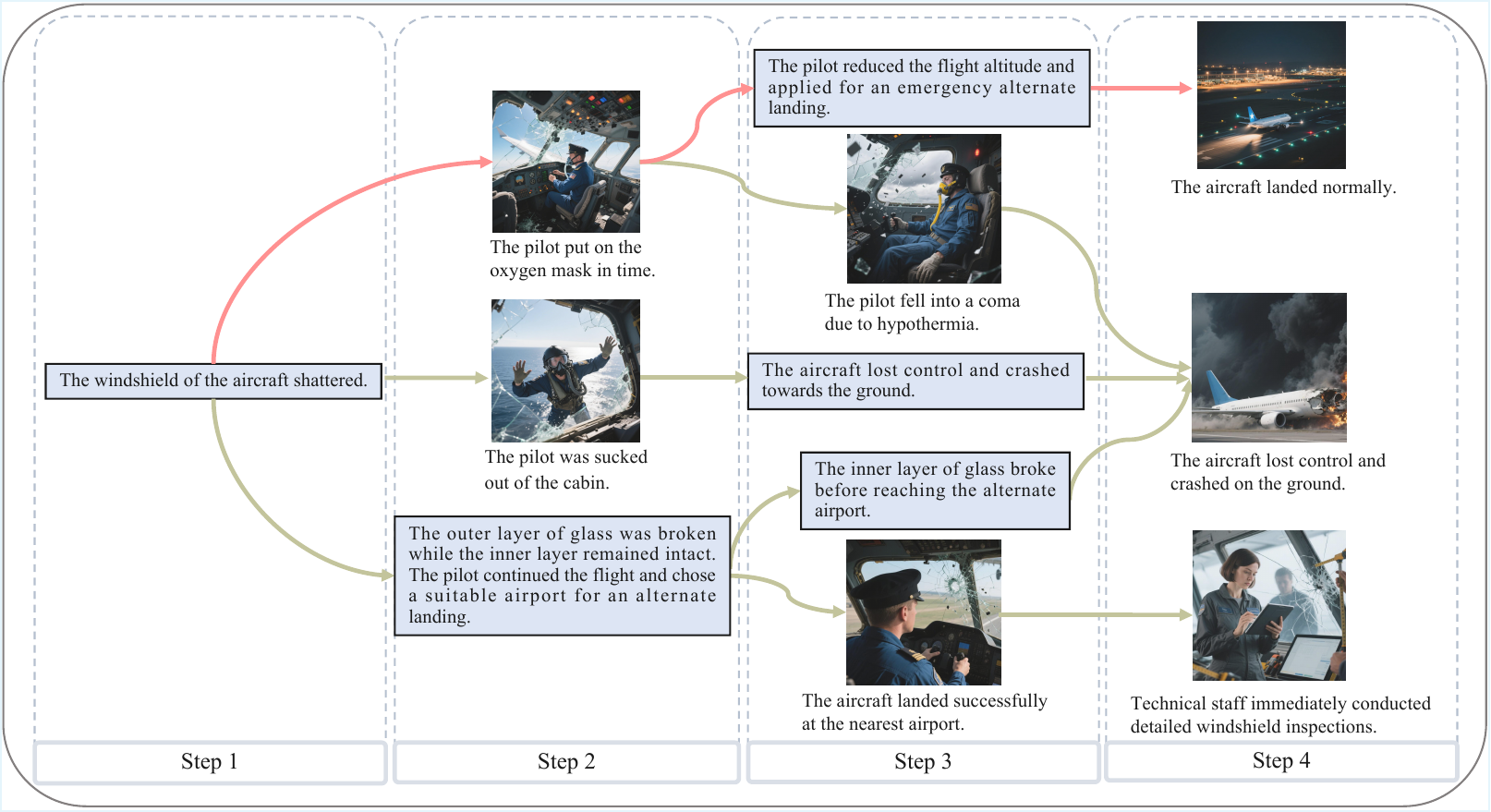}
\caption{The simulation and deduction results of the proposed WLDS in the aircraft's emergency instance. The red part represents the world line selected by the user.}
\label{fig_4}
\end{figure*}

If $\alpha_{\max}(s) < \delta_{\mathrm{align}}$, a new candidate image $\hat{I} \sim p_{\varphi}(\cdot \mid s, \mathcal{K}_B)$ is generated using a text-to-image model and added to the extended candidate set $\mathcal{I}^{+}(s)$.  
This ensures that events without suitable existing images can still be visually represented.

The final keyframe selection is:
\begin{equation}
I^{*}(s) =
\begin{cases}
\arg\max\limits_{I \in \mathcal{I}^{+}(s)} \alpha(s, I), & \text{if } \max\limits_{I \in \mathcal{I}^{+}(s)} \alpha(s, I) \ge \delta_{\mathrm{align}}, \\[0.6em]
\varnothing, & \text{otherwise}.
\end{cases}
\label{Eq 8}
\end{equation}
where $\mathcal{I}^{+}(s)$ is the extended candidate set.

The visualized world line can be formalized as: 
\begin{equation}
V = \big[(s_0, I^{*}(s_0)),\; (s_1, I^{*}(s_1)),\; \dots,\; (s_Q, I^{*}(s_Q))\big].
\label{Eq 9}
\end{equation}

Finally, the visualization sequence \(\mathcal{V} = \{ V_0, V_1, \ldots, V_Q \}\) is constructed by synchronizing the filtered keyframes with the text descriptions of events in \(\mathcal{W}^*\). 
This multimodal representation enables intuitive visualization of the world line’s evolution. The demo of multiple world lines output by WLDS is shown in Figure.\ref{fig_4}.

\subsection{Design of Evaluation Metrics}
To comprehensively evaluate the performance of WLDS, we consider two complementary dimensions: \textit{factual consistency} (FC) and \textit{logical consistency} (LC).

(1) Factual Consistency (FC): Let $\mathcal{E} = \{e_1, e_2, \dots, e_T\}$ be the set of events in a world line, and $\phi_{\mathrm{fact}}(e) \in \{0,1\}$ indicate whether event $e$ aligns with the domain knowledge base.  
FC is defined as:
\begin{equation}
    \mathrm{FC} = \frac{\left| \{ e \in \mathcal{E} \mid \phi_{\mathrm{fact}}(e) = 1 \} \right|}{|\mathcal{E}|},
\label{Eq 10}
\end{equation}
where the numerator counts factually consistent events, and the denominator is the total number of events.

This metric measures the proportion of events that are factually correct, reflecting the system’s accuracy and reliability under knowledge constraints.

(2) Logical Consistency (LC): Let $\mathcal{P} = \{ (e_i,e_{i+1}) \}_{i=1}^{T-1}$ be the set of adjacent event pairs, and $\psi_{\mathrm{logic}}(e_i,e_j) \in \{0,1\}$ indicate whether the pair is logically valid.  
LC is defined as:
\begin{equation}
    \mathrm{LC} = \frac{\left| \{ (e_i,e_j) \in \mathcal{P} \mid \psi_{\mathrm{logic}}(e_i,e_j) = 1 \} \right|}{|\mathcal{P}|}.
\label{Eq 11}
\end{equation}
LC evaluates the stability and coherence of the reasoning chain across multi-step deductions. 

Both FC and LC take values in $[0,1]$, with higher scores indicating better factual alignment and logical coherence. This dual-metric design provides a robust and interpretable basis for quantitative performance analysis. The entire workflow of WLDS is summarized in Algorithm 1.

\begin{algorithm}[h]
\caption{Workflow of the LMs-driven World Line Divergence System (WLDS)}
\label{alg:wlds}
\renewcommand{\algorithmicrequire}{\textbf{Input:}}
\renewcommand{\algorithmicensure}{\textbf{Output:}}
\begin{algorithmic}[1]
\REQUIRE Accident dataset $\mathcal{D}_{\mathrm{acc}}$, domain-specific knowledge base $\mathcal{K}_B$, prompt template $\pi$
\ENSURE Visualized world line $V$, evaluation metrics (FC, LC)

\STATE Generate domain-specific emergency instances using Eq.\ref{Eq 1}, and construct transferred dataset $\mathcal{D}_{\mathrm{trans}}$.
\STATE Select initial event $s_0 \in \mathcal{D}_{\mathrm{trans}}$, generate candidate scenarios under different temperature settings by Eq.\ref{Eq 2} and Eq.\ref{Eq 3}, form candidate set $\mathcal{S}$ and initial world line $W$.
\STATE For each event, compute factual consistency score by $\phi_{\mathrm{fact}}(s, f(s))$. If $\phi_{\mathrm{fact}}(s, f(s)) < \delta_{\mathrm{fact}}$, revise the event using Eq.\ref{Eq 4}.  
\STATE For each adjacent event pair, check logical validity by $\psi_{\mathrm{logic}}(s_i, s_j)$. If invalid, regenerate the subsequent event using Eq.\ref{Eq 5}, and update the world line as in Eq.\ref{Eq 6}.
\STATE For each event, compute maximum alignment score by Eq.\ref{Eq 7}. If below $\delta_{\mathrm{align}}$, generate additional image using a text-to-image model and select final keyframe by Eq.\ref{Eq 8}. Construct visualized world line $V$ using Eq.\ref{Eq 9}.
\STATE Compute factual consistency (FC) and logical consistency (LC) using Eq.\ref{Eq 10} and Eq.\ref{Eq 11}.
\RETURN $V$, (FC, LC)
\end{algorithmic}
\end{algorithm}
\section{Experiment}
In our experiments, we aim to: 
(1) evaluate whether WLDS can achieve high-precision and high-fidelity simulation and deduction in specific domains represented by carrier-based aircraft and urban rail transit,
(2) assess the effectiveness of the factual calibration mechanism in real-time calibration of factually deviation content by relying on domain knowledge bases, 
(3) evaluate the effectiveness of the logical calibration mechanism in calibration logically deviation content by analyzing causal relationships,
(4) validate the effectiveness of the EID benchmark dataset in the performance evaluation and optimization of emergency instances deduction models. 
All experiments were run on two A6000 GPUs.
The professional knowledge base employed by the factual calibration mechanism is constructed based on professional books and instruction manual. The code and data for the proposed method are provided for research purposes. \footnote{Code is included in the supplemental material and will be released upon the paper acceptance.}

\subsection{Introduction of EID benchmark dataset}
\begin{figure}[!t]
\centering
\includegraphics[width=\linewidth]{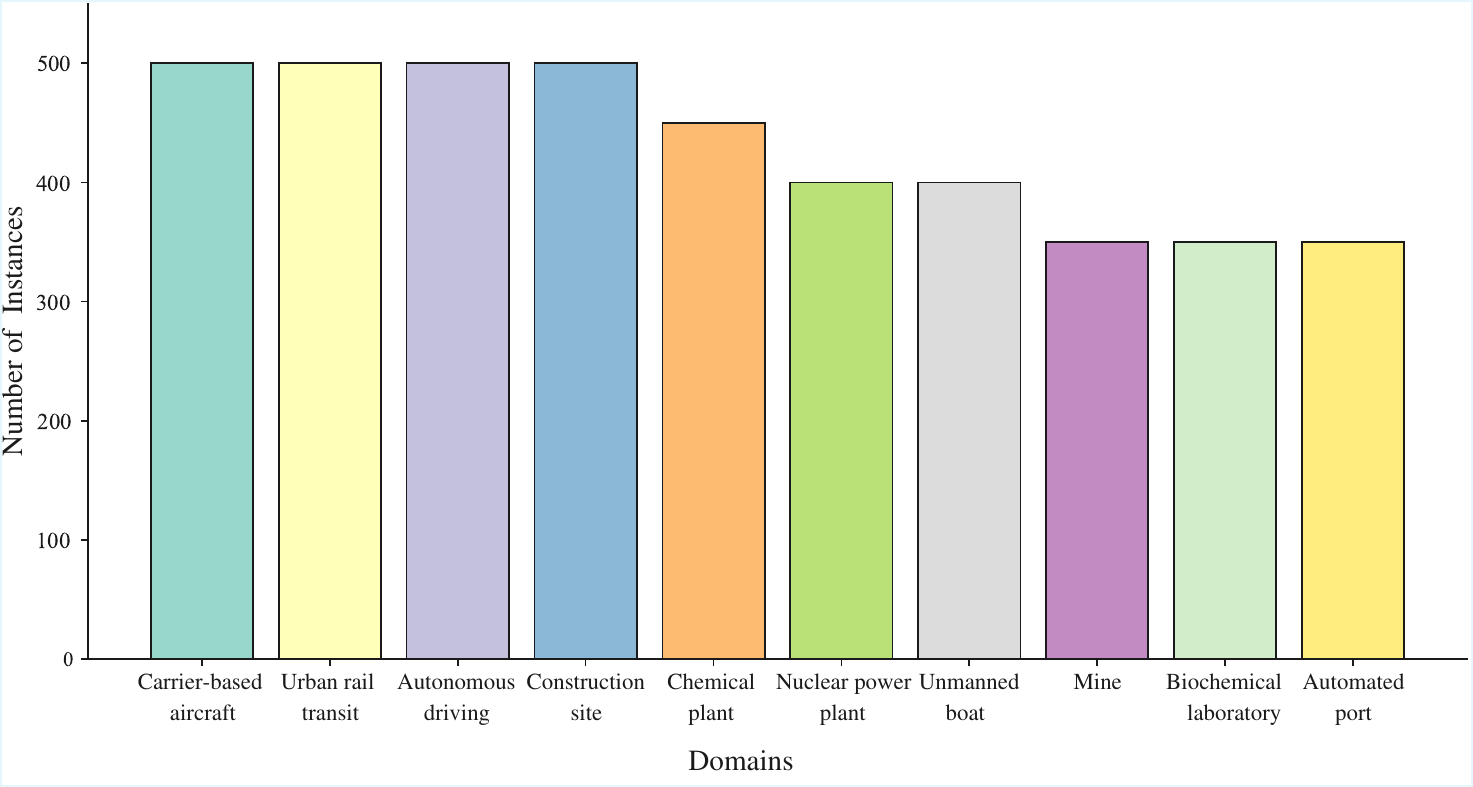}
\caption{Statistical distribution of the 10 sub-datasets of EID benchmark dataset.}
\label{fig_5}
\end{figure}

\begin{table*}[!ht]
  \centering
  \caption{Quantitative comparison between WLDS and baseline models in the 10 sub-datasets of EID benchmark dataset. The evaluated models include WLDS+H (WLDS combined with Hunyuan-Turbos), WLDS+G (WLDS combined with GLM-4-Plus), and WLDS+Q (WLDS combined with Qwen-Max). The performance of each model is measured in terms of FC, LC and the scenario prediction accuracy on the sub-datasets of EID benchmark dataset.}
  \begingroup
  \tiny
  \setlength{\tabcolsep}{1pt}      
  \renewcommand{\arraystretch}{1.2}

  \makebox[\textwidth][c]{%
    \resizebox{1\textwidth}{!}{
      \begin{minipage}[t]{0.485\textwidth}
        \centering
        \begin{tabular}{c c c c c} 
          \toprule
          \textbf{Datasets} & \textbf{Methods} & \textbf{FC} & \textbf{LC} & \textbf{EID} \\
          \midrule
          \multirow{6}{*}{Carrier-based aircraft}
            & Hunyuan & 85.11\% & 82.92\% & 84.20\% \\
            & WLDS+H & \textbf{90.42\%} & \textbf{89.58\%} & 88.60\% \\
            & GLM & 85.42\% & 83.75\% & 85.40\% \\
            & WLDS+G & 89.33\% & 87.43\% & \textbf{89.20\%} \\
            & Qwen & 86.25\% & 82.08\% & 84.80\% \\
            & WLDS+Q & 90.29\% & 88.24\% & 88.60\% \\   
          \midrule
          \multirow{6}{*}{Urban rail transit} 
            & Hunyuan & 86.67\% & 83.33\% & 82.40\% \\
            & WLDS+H & \textbf{93.75\%} & \textbf{91.67\%} & \textbf{90.20\%} \\
            & GLM & 87.08\% & 85.42\% & 81.20\% \\
            & WLDS+G & 92.29\% & 90.62\% & 89.60\% \\
            & Qwen & 88.33\% & 84.58\% & 80.60\% \\
            & WLDS+Q & 91.43\% & 91.18\% & 88.60\% \\
          \midrule
          \multirow{6}{*}{Autonomous driving} 
            & Hunyuan & 90.00\% & 88.75\% & 85.20\% \\
            & WLDS+H & \textbf{94.17\%} & \textbf{93.33\%} & \textbf{90.80\%} \\
            & GLM & 89.19\% & 87.50\% & 84.00\% \\
            & WLDS+G & 93.75\% & 91.17\% & 87.80\% \\
            & Qwen & 90.32\% & 87.88\% & 85.80\% \\
            & WLDS+Q & 93.55\% & 91.29\% & 89.80\% \\
          \midrule
          \multirow{6}{*}{Construction site} 
            & Hunyuan & 87.50\% & 84.17\% & 82.22\% \\
            & WLDS+H & \textbf{92.92\%} & \textbf{91.25\%} & \textbf{89.40\%} \\
            & GLM & 86.25\% & 84.58\% & 81.56\% \\
            & WLDS+G & 92.57\% & 90.91\% & 88.22\% \\
            & Qwen & 87.14\% & 83.75\% & 80.89\% \\
            & WLDS+Q & 91.84\% & 90.62\% & 89.11\% \\
          \midrule
          \multirow{6}{*}{Chemical plant} 
            & Hunyuan & 87.10\% & 85.00\% & 82.75\% \\
            & WLDS+H & 91.25\% & \textbf{90.42\%} & 90.75\% \\
            & GLM & 88.33\% & 84.83\% & 83.00\% \\
            & WLDS+G & 91.98\% & 89.58\% & \textbf{91.50\%} \\
            & Qwen & 88.75\% & 85.96\% & 82.75\% \\
            & WLDS+Q & \textbf{92.56\%} & 90.32\% & 89.25\% \\    
          \bottomrule
        \end{tabular}
      \end{minipage}
      \hspace{4pt}\vrule width 0.4pt\hspace{4pt}%
      \begin{minipage}[t]{0.485\textwidth}
        \centering
        \begin{tabular}{c c c c c}
          \toprule
          \textbf{Datasets} & \textbf{Methods} & \textbf{FC} & \textbf{LC} & \textbf{EID} \\
          \midrule
          \multirow{6}{*}{Nuclear power plant} 
            & Hunyuan & 85.83\% & 85.71\% & 82.00\% \\
            & WLDS+H & \textbf{90.32\%} & 90.00\% & \textbf{89.14\%} \\
            & GLM & 85.71\% & 86.11\% & 83.71\% \\
            & WLDS+G & 89.97\% & \textbf{90.62\%} & 87.43\% \\
            & Qwen & 85.42\% & 84.58\% & 82.86\% \\
            & WLDS+Q & 89.43\% & 89.57\% & 88.29\% \\       
          \midrule
          \multirow{6}{*}{Unmanned boat} 
            & Hunyuan & 88.57\% & 87.21\% & 86.60\% \\
            & WLDS+H & 93.33\% & 91.41\% & 90.60\% \\
            & GLM & 89.58\% & 87.10\% & 86.40\% \\
            & WLDS+G & 92.29\% & 91.14\% & \textbf{91.20\%} \\
            & Qwen & 88.33\% & 86.67\% & 87.20\% \\
            & WLDS+Q & \textbf{93.57\%} & \textbf{92.50\%} & 90.40\% \\       
          \midrule
          \multirow{6}{*}{Mine} 
            & Hunyuan & 87.08\% & 82.92\% & 85.71\% \\
            & WLDS+H & \textbf{92.08\%} & \textbf{90.32\%} & 88.57\% \\
            & GLM & 85.83\% & 84.17\% & 84.29\% \\
            & WLDS+G & 91.18\% & \textbf{90.32\%} & 88.00\% \\
            & Qwen & 86.67\% & 83.33\% & 85.43\% \\
            & WLDS+Q & 90.29\% & 89.86\% & \textbf{89.43\%} \\        
          \midrule
          \multirow{6}{*}{Biochemical laboratory}
            & Hunyuan & 87.33\% & 86.01\% & 86.00\% \\
            & WLDS+H & \textbf{91.67\%} & 90.91\% & \textbf{89.75\%} \\
            & GLM & 86.67\% & 87.50\% & 86.25\% \\
            & WLDS+G & 91.14\% & \textbf{91.71\%} & \textbf{89.75\%} \\
            & Qwen & 85.42\% & 84.07\% & 84.75\% \\
            & WLDS+Q & 89.74\% & 88.29\% & 88.75\% \\
          \midrule
          \multirow{6}{*}{Automated port} 
            & Hunyuan & 89.07\% & 87.54\% & 87.43\% \\
            & WLDS+H & 91.55\% & \textbf{92.92\%} & 90.85\% \\
            & GLM & 88.57\% & 87.92\% & 86.29\% \\
            & WLDS+G & 91.71\% & 90.83\% & 90.29\% \\
            & Qwen & 89.58\% & 86.58\% & 87.71\% \\
            & WLDS+Q & \textbf{92.05\%} & 91.88\% & \textbf{91.43\%} \\  
          \bottomrule
        \end{tabular}
      \end{minipage}
    }
  }
  \endgroup
  \label{tab:performance_comparison} 
\end{table*}

To enhance the multi-step deduction capability of reasoning models for emergency instances, we propose Emergency Instances Deduction (EID) benchmark dataset which is a benchmark dataset focused on multi-step emergency instances deduction. 
As shown in Figure.\ref{fig_5}, EID benchmark dataset contains 4,300 high-quality three-step deduction data entries of emergency instances, covering 10 sub-datasets such as EID-Carrier-based aircraft and EID-Urban rail transit.
Each data entry starts from an initial emergency instance and forms 14 diverse branch scenarios through a three-stage structured deduction. 
The labels annotated by humans include the most probable scenario, as well as the probability and loss severity of each world line.
This dataset fills the gap in the dynamic modeling of emergency instances deduction in existing benchmark datasets.
Furthermore, it provides high-quality data for training and evaluating the multi-step reasoning capability, scenario prediction accuracy, and domain-adaptive reasoning performance of emergency instances deducing models.  

\subsection{Quantitative Analysis of WLDS}
In the quantitative experiment, Fact Consistency (FC) and Logical Consistency (LC) were employed to quantitatively verify the effectiveness of WLDS from the two dimensions of fact and logic.
Meanwhile, comparative experiments between WLDS and three LMs (Hunyuan-Turbos, GLM-4-Plus, and Qwen-Max), are conducted to verify the supporting role of the benchmark dataset EID in the performance evaluation of reasoning models.
All models are invoked through APIs.


The experimental results are shown in Table.\ref{tab:performance_comparison}. 
In terms of FC and LC, WLDS achieves an average FC of 91.75\% and an average LC of 90.66\% across all tested domains, significantly outperforming the other comparison methods, demonstrating its effectiveness in emergency instances deduction.
Specifically, in the domain of urban rail transit, WLDS demonstrates significant improvements over the baseline model.
The FC and LC of Hunyuan-Turbos are 86.67\% and 83.33\% respectively.  In contrast, when combined with Hunyuan-Turbos, WLDS can achieve FC and LC of 93.75\% and 91.67\%, representing relative improvements of 7.08\% and 8.34\%.
This improvement indicates that WLDS can effectively enhance factual consistency and logical consistency during the deduction process in complex emergency instances deduction. 
Furthermore, WLDS achieved the highest scenario prediction accuracy of 91.50\% in EID-Chemical plant, a notable improvement of 8.50\% compared to the baseline model GLM. This result not only validates WLDS’s robustness in multi-step emergency instances deduction but also highlights the supporting role of the EID benchmark dataset in performance evaluation. 
Additionally, WLDS’s FC of 94.17\% and LC of 93.33\% in the autonomous driving domain represent the best performance across all tested domains. 
In other domains, WLDS maintains an FC above 90\%, further proving its stability and reliability in emergency instances deduction. The outstanding performance of WLDS in various specific domains reflects its powerful capabilities in emergency instances deduction, providing a new approach for the simulation and deduction in various domains.


\subsection{Performance of WLDS}  

\begin{figure*}[!ht]
\centering
\includegraphics[width=\linewidth]{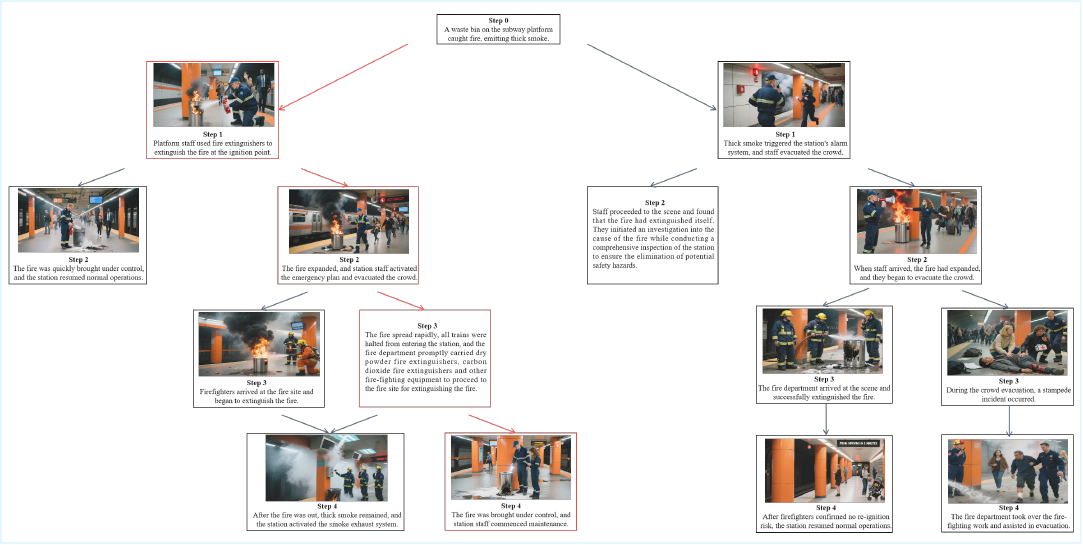}
\caption{The results of emergency instances deduction by WLDS in the urban rail transit domain. The red path denotes the user-selected world line.}
\label{fig_6}
\end{figure*}

Figure.\ref{fig_6} presents the simulation and deduction results of WLDS in the emergency insatances of the urban rail transit domain.
WLDS takes "A waste bin on the subway platform caught fire, emitting thick smoke", as the initial emergency instance, and generates 7 world lines, sequentially named World Line 0 to World Line 6.
Under the same initial emergency instance, each world line incorporates controlled randomness and diversity, generating multi-stage evolutionary processes ranging from mildly controllable fire situations to high-risk states involving crowd stampedes.

\begin{figure}[!ht]
  \centering
  \subfloat[Visualization of the probability and loss severity of the WLDS-generated world lines.\label{fig:12a}]{
    \includegraphics[width=0.65\linewidth]{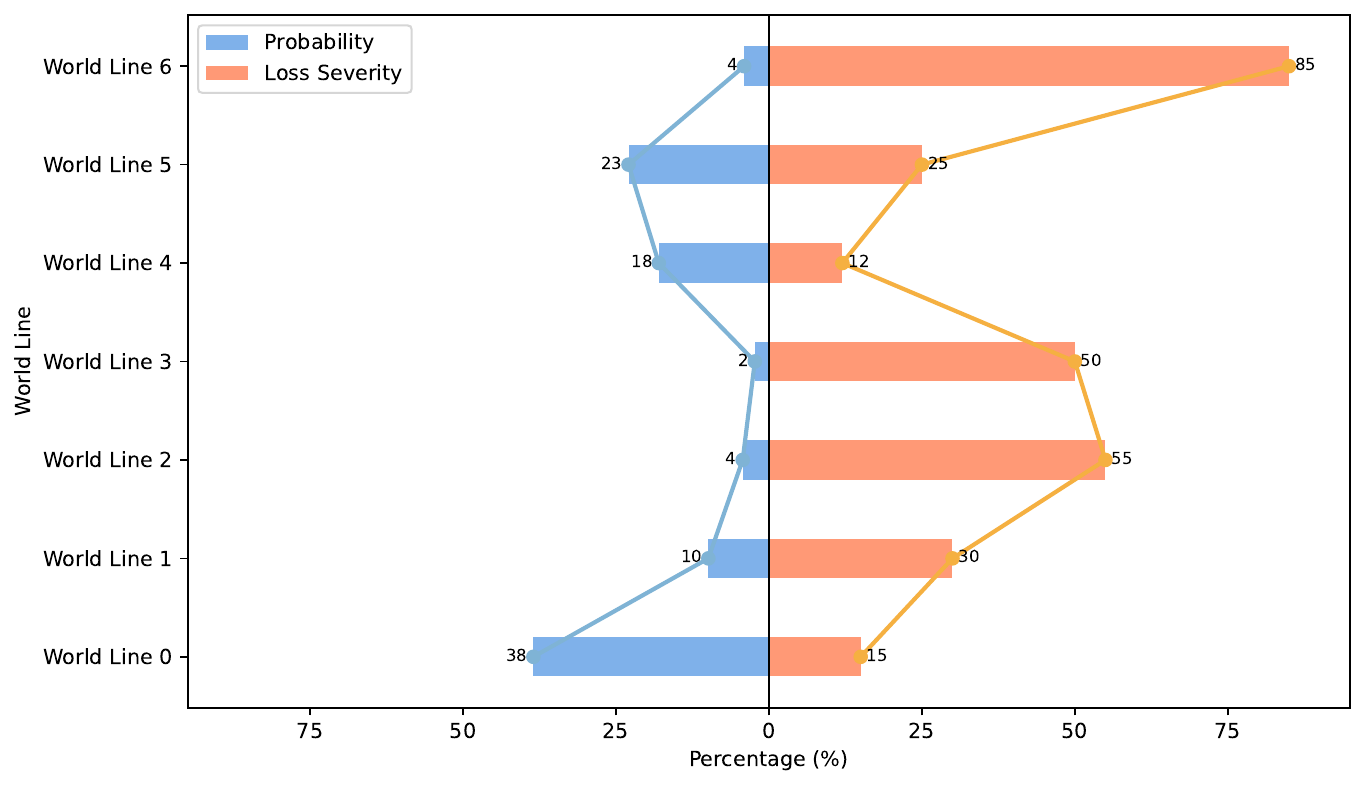}
  }\hfill
  \subfloat[Knowledge graph constructed from the seven world lines.\label{fig:12b}]{
    \includegraphics[width=0.5\linewidth]{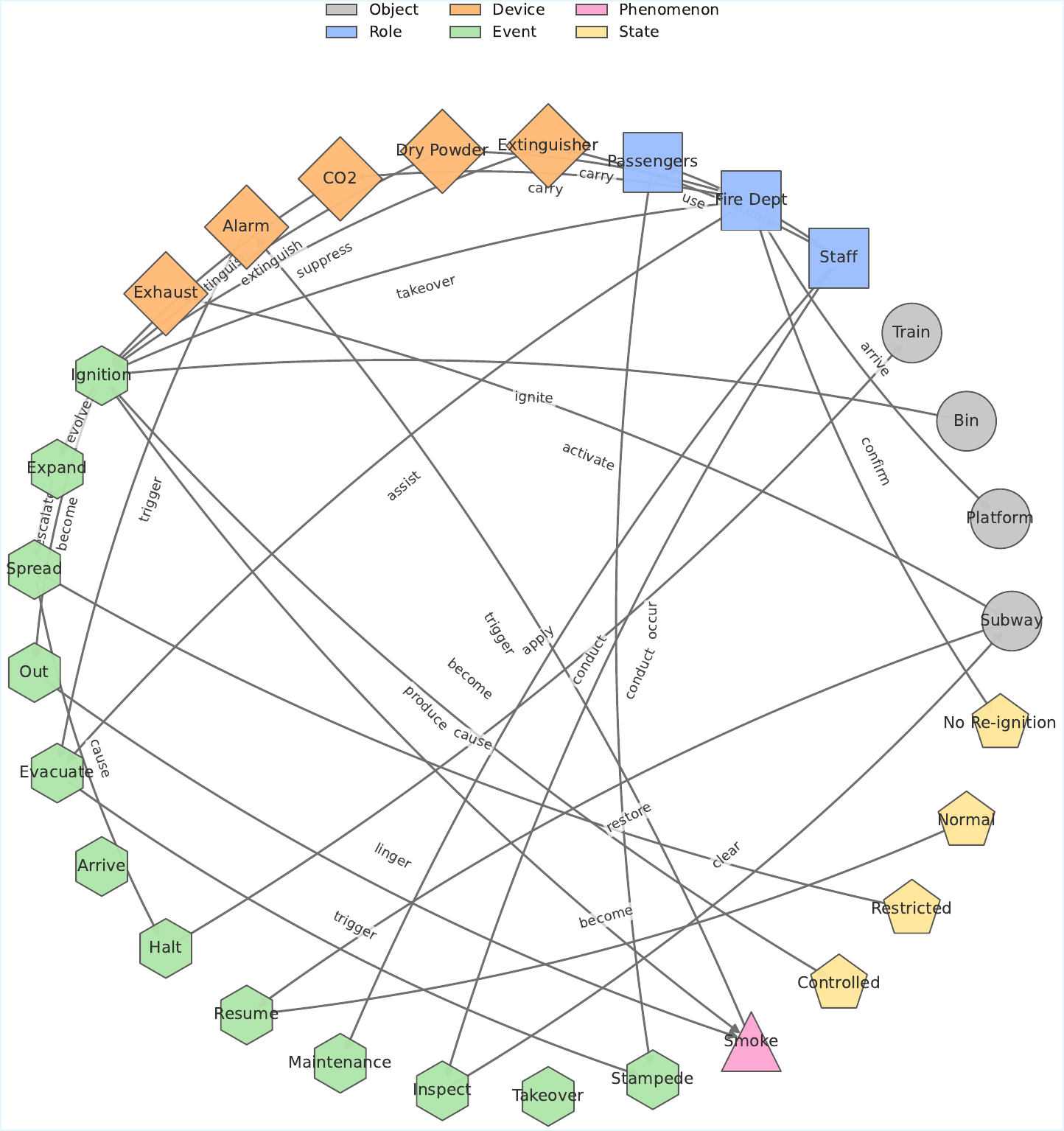}
  }
  \caption{WLDS outputs: (a) probability and loss severity across the seven generated world lines; (b) the corresponding knowledge graph.}
  \label{fig:12}
\end{figure}

From the perspective of event evolution logic, the 7 world lines depicted in Figure.\ref{fig_6} exhibit a progressive evolutionary feature from low risk to high risk, with distinct differences in response strategies: 
World Line 0 represents a typical low-risk scenario, where the fire was effectively extinguished by staff using fire extinguishers at the initial stage, enabling the rapid restoration of normal operations. 
World Line 1 falls into a low-to-moderate risk scenario which shows that the initial fire suppression failed to fully control the fire, leading to the activation of emergency plans and passenger evacuation.
After fire department intervention extinguished the fire, the smoke exhaust system was activated for post-incident handling. 
Both World Line 2 and World Line 3 represent moderate-risk scenarios, where the fire spread rapidly, forcing trains to stop entering the station. 
The fire department conducted extinguishing operations using equipment such as dry powder and carbon dioxide fire extinguishers. 
The difference lies in that World Line 2 resumed operations directly after fire extinguishment and smoke exhaust, while World Line 3 implemented equipment inspection and platform maintenance after fire control to optimize subsequent operational safety. 
World Line 4 shows a risk-recession path which shows that after heavy smoke triggered the fire alarm system and passengers were evacuated, staff found that the fire had extinguished itself, and then shifted to fire cause investigation and full-station safety inspection to eliminate potential hazards. 
World Line 5 belongs to a moderate-to-high risk scenario, where fire expansion prompted simultaneous evacuation and fire-fighting actions, with operations resumed after extinguishment and confirmation of no re-ignition risk. 
World Line 6 represents the highest-risk scenario which shows that during a fire development process similar to that of World Line 5, a stampede occurred during the evacuation phase. 
In addition to fire extinguishment, the fire department also assisted in crowd management and safe evacuation, reflecting the need for multi-departmental emergency coordination in complex disaster scenarios.


From the perspectives of factual accuracy and operational standardization, each step complies with subway safety regulations. 
Direct fire suppression in low-risk scenarios aligns with the principle of on-site rapid disposal for initial fires.
The procedures such as operation suspension, evacuation, smoke exhaust, and equipment inspection in moderate-to-high risk scenarios are consistent with subway operational safety standards.
And the crowd control and medical assistance in stampede incidents meet the emergency response requirements for sudden crowd accidents.

\begin{figure}[!b]
\centering
\includegraphics[width=\linewidth]{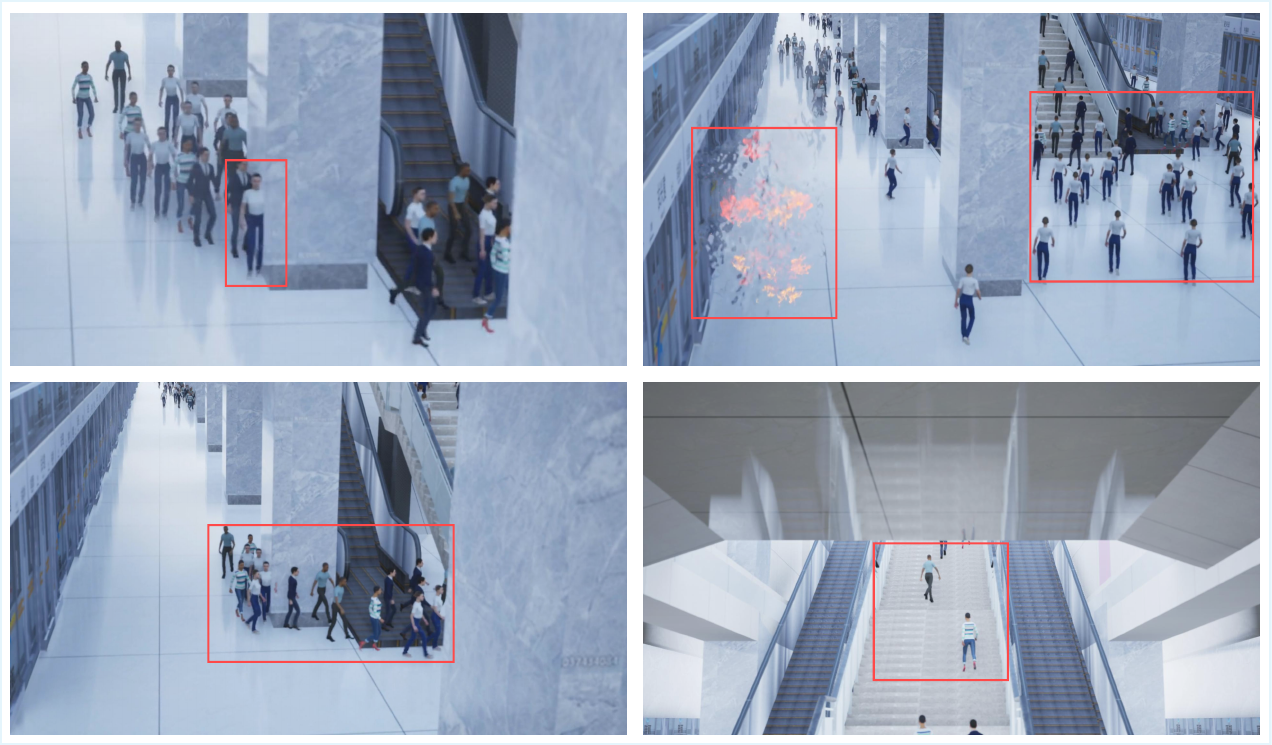}
\caption{Partial screenshots of a conventional urban rail transit simulation system.}
\label{fig_7}
\end{figure}

In addition, for the 7 world lines deduced by WLDS, we employed the GLM4-9B model, trained on the EID benchmark dataset, to analyze each world line from the dimensions of probability and loss severity, and visualized the results to assist the user in understanding each world line.
As shown in Figure.\ref{fig:12a}, the 7 world lines generated by WLDS exhibit different probability and loss severity.
This demonstrates that WLDS not only generates diverse deduction directions but also presents a broader range of risk evolution patterns while maintaining high fidelity.

As shown in Figure.\ref{fig:12b}, WLDS generates a knowledge graph representation of the 7 deduced world lines. 
The nodes cover multiple categories such as objects, roles, devices, events, phenomena, and states, while the edges represent their logical and causal relationships. 
Through this knowledge graph, the causal chains and interactions of emergency instances under different deduction paths can be more intuitively revealed, thereby helping users understand the complex logical dynamics in multi-path deductions.

\begin{table}[!b]
  \centering
  \caption{Results of the ablation study, comparing WLDS performance after removing logical calibration or factual calibration, demonstrating the contribution of each core module.}
  \renewcommand{\arraystretch}{1.2} 
  \begin{tabular}{c c c c} 
    \toprule
    \textbf{Methods} & \textbf{FC} & \textbf{LC} & \textbf{EID} \\
    \midrule
    Hunyuan & 86.67\% & 83.33\% & 82.40\% \\
    Hunyuan + Factual calibration & 90.42\% & 87.50\% & 87.40\% \\
    Hunyuan + Logical calibration & 88.75\% & 89.58\% & 84.60\% \\
    WLDS & 93.75\% & 91.67\% & 90.20\% \\
    \bottomrule
  \end{tabular}
  \label{tab:single_method_comparison}
\end{table}

Figure.\ref{fig_7} illustrates the performance of the existing traditional urban rail transit simulation system.
From the figure, the shortcomings can be clearly observed: interpenetration between human models, visually unrealistic flame effects, duplicated avatars leading to a lack of character distinctiveness, and human postures that deviate from physical laws. 
These defects not only weaken the understanding of crowd intentions but also reduce the reliability of risk assessment and causal reasoning. In contrast, WLDS offers significant advantages in the following three aspects:  
(1) Factual and logical constraints: Through dual factual and logical calibration, WLDS confines deduction to states and processes consistent with domain knowledge, reducing the non-physical or program-logic errors frequently seen in the traditional simulation.  
(2) Diverse world-line generation: By leveraging LMs for emergency instances deduction, WLDS generates multiple reasonable world lines under knowledge constraints, while maintaining both randomness and diversity. 
(3) Semantically aligned keyframe visualization: By employing text–image alignment assessment and image generation, WLDS selects or synthesizes keyframe images consistent with the narrative of deduction, making high expressiveness the primary objective of simulation and deduction.
In summary, the 7 world lines deduced by WLDS cover the complete risk spectrum, ranging from slightly controllable fires to incidents involving personal injuries.
These results highlight WLDS's capability for high-fidelity and diversified scenario generation and offer a visualized and verifiable experimental basis for risk assessment and decision support in the urban rail transit domain.

\subsection{Ablation Study: Importance of Dual Calibration Mechanism}

\begin{figure}[!b]
\centering
\includegraphics[width=\linewidth]{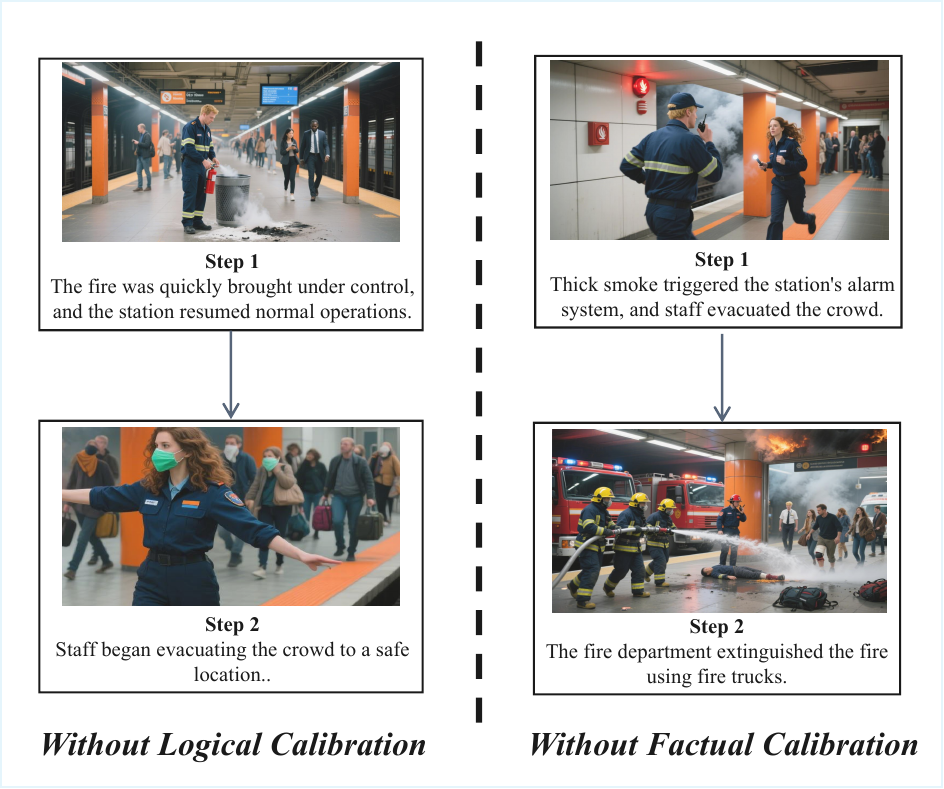}
\caption{Comparative deduction results of WLDS in the urban rail transit domain without logical calibration and without factual calibration.}
\label{fig_9}
\end{figure}

WLDS introduces factual calibration and logical calibration to address factual deviation and logical deviation.
To quantitatively evaluate the specific contributions of these two mechanisms to the performance of WLDS, this section conducts the ablation study.  
We removed factual calibration and logical calibration respectively to perform deduction on emergency instances.  
All initial scenarios of the experiments are derived from emergency instances in the domain of urban rail transit.  
The quantitative results are presented in Table.\ref{tab:single_method_comparison}.

The results indicate that the WLDS model achieves a factual consistency of 93.75\%, a logical consistency of 91.67\%, and an accuracy of 90.2\% on the EID benchmark dataset, all of which outperform the performance of models using only the factual calibration mechanism or the logical calibration mechanism individually. 
This validates that the dual calibration mechanism not only ensures the factual reliability of individual events through factual calibration but also maintains the causal continuity between consecutive events via logical calibration, thereby collectively enabling WLDS to overcome the hallucination issues of LMs and achieve high-precision and high-fidelity simulation and deduction of emergency incidents.

\newcolumntype{L}[1]{>{\raggedright\arraybackslash}m{#1}}
\begin{table*}[!b]
  \centering
  \caption{Evaluation criteria used by domain experts for scoring WLDS deduction performance.}
  \label{tab:wlds-criteria}
  \scriptsize                       
  \setlength{\tabcolsep}{4pt}       
  \renewcommand{\arraystretch}{1.06}
  \begin{tabular}{@{}L{0.13\textwidth} L{0.40\textwidth} L{0.42\textwidth}@{}}
    \toprule
    \textbf{Evaluation dimension} & \textbf{Refined assessment indicators} & \textbf{Scoring criteria (0--5 points)} \\
    \midrule
    \textbf{Diversity} &
    (1) Whether the world lines cover the full spectrum of risk levels, from minor faults to severe accidents. \newline
    (2) Whether the scenarios include both single-factor special conditions and multi-factor coupled situations. &
    \textbf{5 points}: The world lines cover no fewer than three risk levels and at least two types of multi-factor coupled scenarios. \newline
    \textbf{0 points}: The world lines repeatedly present low-risk, low-complexity scenarios of the same type. \\
    \midrule
    \textbf{Precision} &
    (1) Whether the generated content conforms to physical principles and domain-specific knowledge. \newline
    (2) Whether standardized, domain-specific terminology is used consistently. &
    \textbf{5 points}: The generated content aligns with domain knowledge, and terminology is appropriate and consistent. \newline
    \textbf{0 points}: The generated content contradicts fundamental domain knowledge or employs incorrect terminology. \\
    \midrule
    \textbf{Rigor} &
    (1) Whether the logical progression in the deduction is coherent and consistent. \newline
    (2) Whether the operational procedure matches real-world practice. &
    \textbf{5 points}: The world lines are logically coherent, and every step strictly conforms to real-world operational procedures. \newline
    \textbf{0 points}: The world lines are logically disordered, with many steps diverging from real-world operational procedures. \\
    \midrule
    \textbf{Feasibility} &
    (1) Whether the deduced emergency response plans comply with domain-specific regulations. \newline
    (2) Whether the plans have practical value in real-world scenarios. &
    \textbf{5 points}: The generated content adheres to domain regulations and can be directly applied in real-world scenarios. \newline
    \textbf{0 points}: The generated content violates domain regulations or lacks potential for application. \\
    \bottomrule
  \end{tabular}
  \label{tab3}
\end{table*}

Figure.\ref{fig_9} further demonstrates the significance of the dual calibration mechanism. In the scenario without logical calibration (left figure), the sequence exhibits a distinct logical discontinuity. 
For example, when the event progresses to "the fire was quickly brought under control and the station resumed normal operations", the subsequent step describes "staff beginning to evacuate the crowd to a safe location". 
This constitutes a logical inconsistency, as the evacuation following the station's return to normal operation lacks causal justification. 
In the scenario without factual calibration (right panel), a factual deviation emerges, where the model deduces that "the fire department extinguished the fire using fire trucks".
This is a factual inconsistency because, in the enclosed underground space of a subway platform, fire trucks cannot access the area, making it impossible to use fire trucks for fire extinguishing operations. 
The qualitative observations from Figure.\ref{fig_9} further confirm the critical role of the dual calibration mechanism.

\subsection{Evaluation by Domain Experts}

To further validate the effectiveness of WLDS in practical applications, 20 experts from the relevant domains were invited to conduct an evaluation.
The evaluation dimensions include four aspects: diversity, precision, rigor and feasibility, adopting a 5-point scoring system where 0 point represents the worst performance and 5 points represent the best. 
The relevant evaluation criteria are presented in Table.\ref{tab3}.

\begin{figure}[!b]
\centering
\includegraphics[width=\linewidth]{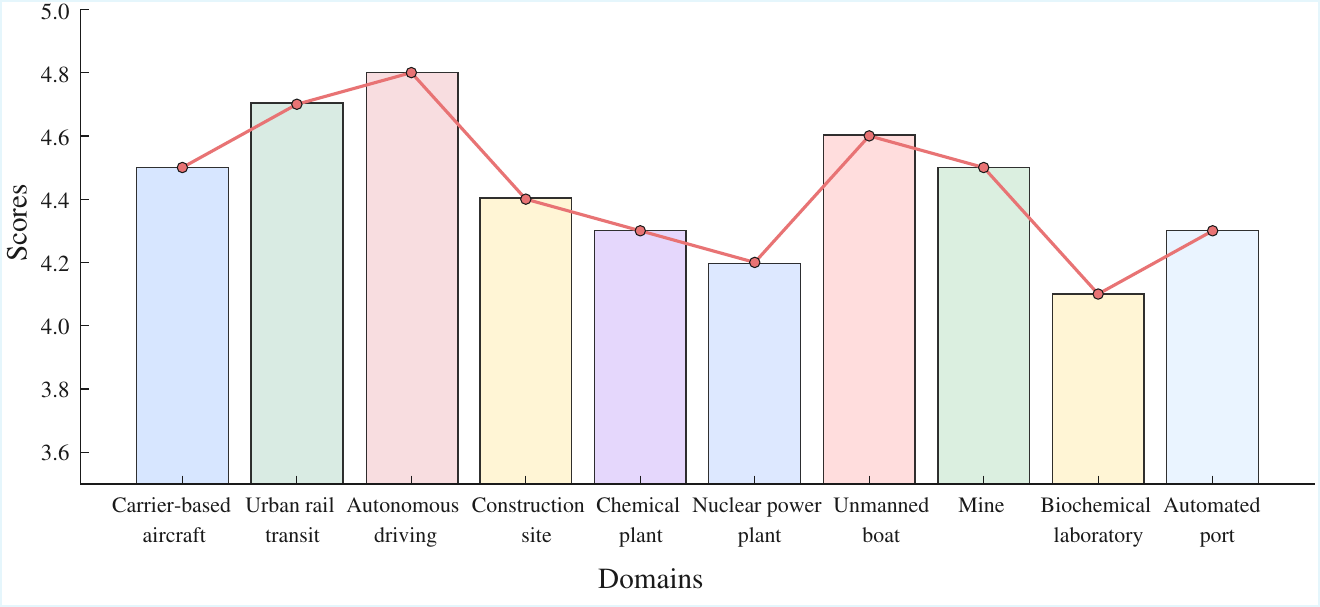}
\caption{Visualization of the expert evaluation scores for WLDS on the 10 sub-datasets of the EID benchmark dataset.}
\label{fig_10}
\end{figure}

Figure.\ref{fig_10} shows that the deduction in different domains of WLDS received consistently high ratings from experts.
Specifically, the autonomous driving domain received the highest average score of 4.8 which reflected strong expert recognition about WLDS's ability to generate diverse, accurate, and logically rigorous emergency instances deduction in this domain.  
The average score reached 4.4, demonstrating that WLDS consistently meets the practical requirements of emergency instances deduction in diverse specific domains.

\section{Discussion}

\textbf{Implications:}  
The proposed WLDS addresses key challenges in emergency instances deduction of specific domains, including the scarcity of emergency instances, limited scenario diversity, and logical inconsistency in simulation outputs. 
By introducing cross-domain knowledge transfer in the instance generation phase, WLDS can produce highly relevant and realistic initial emergency instances deduction even in the absence of sufficient in-domain samples. 
The incorporation of controlled randomness during deduction ensures logical plausibility while enhancing diversity, and the dual calibration mechanism guarantees both factual accuracy and logical coherence. 
Furthermore, the integration of keyframe-based visualization improves interpretability, providing an effective tool for training and evaluating decision-making in response to emergency instances.

\textbf{Limitations and future work:}  
Although WLDS can achieve excellent performance in short-step deduction, logical coherence tends to decrease to some extent as the number of reasoning steps increases.
In addition, While the current image-text integrated visualization approach can intuitively present the evolution of scenarios, it remains insufficient in expressing dynamic interactions, temporal rhythms, and immersive experiences.
In future work, we plan to improve WLDS in two directions:
(1) By incorporating more detailed domain knowledge representations and multi-stage deduction mechanisms, we seek to enhance the accuracy and logical consistency of long-chain deduction tasks.
(2) We plan to incorporate dynamic visualization methods, such as animated scenarios, interactive environments, and video generation, to improve temporal continuity and overall immersion.

\section{Conclusion}
This paper proposes WLDS which is a LMs-driven system for simulation and  deduction of emergency instances.
WLDS generates highly relevant emergency instances via cross-domain knowledge transfer, introduces controlled randomness to produce diverse scenarios, and applies the dual calibration mechanism to ensure both factual accuracy and logical coherence. 
Keyframe-based visualization further enhances interpretability and user understanding.
Experimental results demonstrate that WLDS achieves superior performance in factual consistency, logical consistency, and scenario diversity compared with existing approaches, confirming its effectiveness and applicability in emergency instances deduction.  







\bibliographystyle{elsarticle-num} 
\bibliography{template}            




\end{document}